\documentclass[10pt,twocolumn,letterpaper]{article}

\usepackage{cvpr}              

\usepackage{graphicx}
\usepackage{amsmath}
\usepackage{amssymb}
\usepackage{booktabs}
\usepackage{currfile}

\usepackage[numbers]{natbib}  
\usepackage{enumitem}  
\usepackage{color}
\usepackage{colortbl}
\usepackage{microtype}
\usepackage{nicefrac}
\usepackage{siunitx}
\usepackage[hang,flushmargin]{footmisc}

\usepackage{algorithm}
\usepackage{algpseudocode}

\usepackage[dvipsnames]{xcolor}

\newcommand{\rot}[2][l]{\rotatebox[origin=#1]{90}{#2}}  

\definecolor{turquoise}{cmyk}{0.65,0,0.1,0.3}
\definecolor{purple}{rgb}{0.65,0,0.65}
\definecolor{dark_green}{rgb}{0, 0.5, 0}
\definecolor{orange}{rgb}{0.8, 0.6, 0.2}
\definecolor{dark_orange}{rgb}{0.7, 0.6, 0.3}
\definecolor{red}{rgb}{0.8, 0.2, 0.2}
\definecolor{darkred}{rgb}{0.6, 0.1, 0.05}
\definecolor{blueish}{rgb}{0.0, 0.3, .6}
\definecolor{light_gray}{rgb}{0.7, 0.7, .7}
\definecolor{pink}{rgb}{1, 0, 1}
\definecolor{cyan}{rgb}{0., 1, 1}

\definecolor{checkyes}{rgb}{0.7, 1.0, 0.7}
\definecolor{crossno}{rgb}{1.0, 0.7, 0.7}
\def \y {$\checkmark$\cellcolor{checkyes}}
\def \n {$\times$\cellcolor{crossno}}

\definecolor{tabbestcolor}{rgb}{0.785, 0.851, 0.969}
\def \best {\cellcolor{tabbestcolor!85}}
\def \sbest {\cellcolor{tabbestcolor!30}}

\definecolor{mydarkblue}{rgb}{0,0.08,0.55}

\renewcommand{\paragraph}[1]{\vspace{.2em}\noindent\textbf{#1}.}

\definecolor{cvprblue}{rgb}{0.21,0.49,0.74}
\usepackage[pagebackref,breaklinks,colorlinks,citecolor=cvprblue]{hyperref}

\def\paperID{177} 
\def\confName{3DV\xspace}
\def\confYear{2024\xspace}

\title{Consistent-1-to-3: Consistent Image to 3D View Synthesis \\ via Geometry-aware Diffusion Models}

\author{Jianglong Ye$^{1}$ \qquad Peng Wang$^{2}$\qquad Kejie Li$^{2}$\qquad Yichun Shi$^{2}$ \qquad Heng Wang$^{2}$ \\
  $^{1}$UC San Diego\qquad $^{2}$ByteDance }

\begin{document}

\twocolumn[{
      \vspace{-1em}
      \maketitle
      \vspace{-1em}
      \begin{center}
        \centering
        \vspace{-0.2in}
        \includegraphics[width=\linewidth]{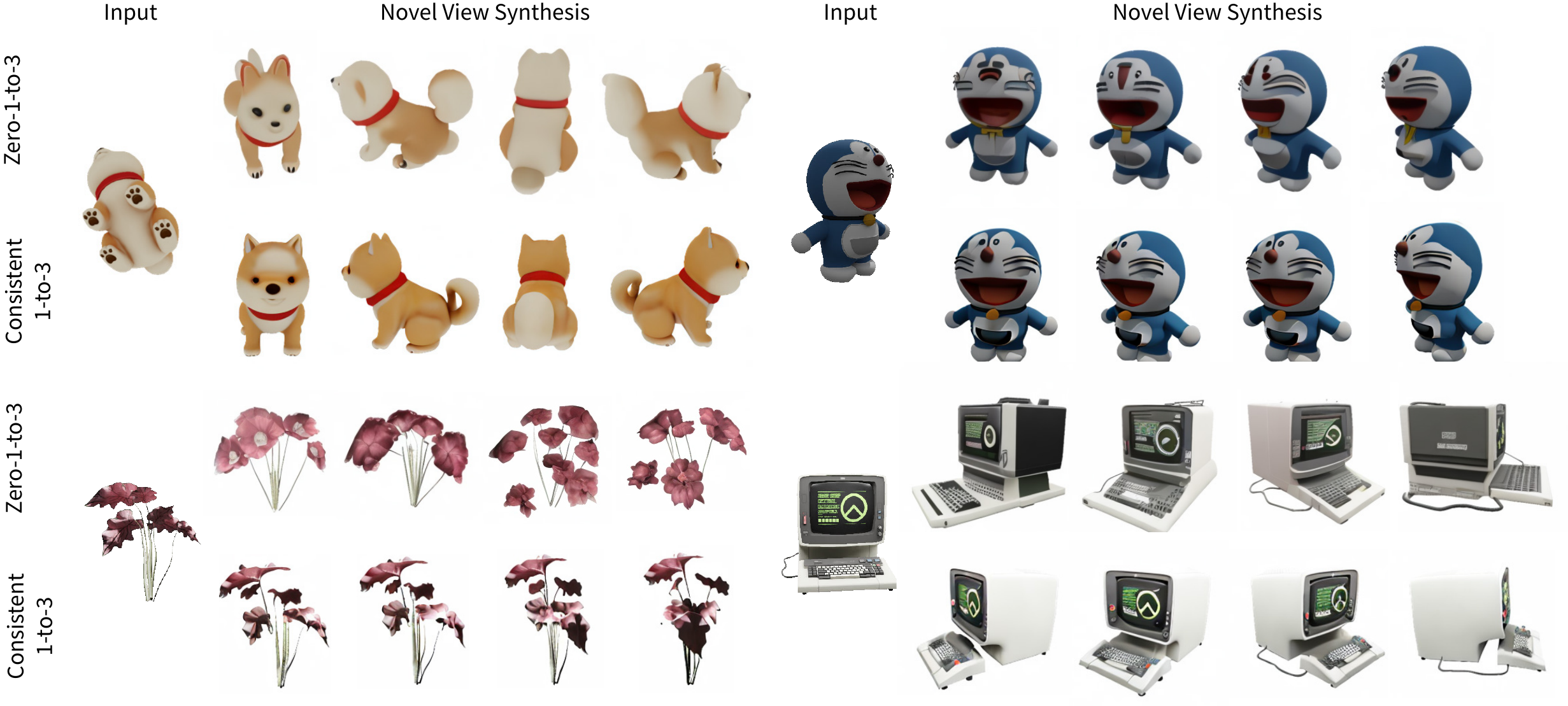}
        \vspace{-0.2in}
        \captionof{figure}{
          \textbf{Consistent-1-to-3} is a novel framework that generates consistent images of any objects from any viewpoint given a single image.
        }
        \label{fig:teaser}
      \end{center}
    }]

\maketitle
\begin{abstract}
Zero-shot novel view synthesis (NVS) from a single image is an essential problem in 3D object understanding. 
While recent approaches that leverage pre-trained generative models can synthesize high-quality novel views from in-the-wild inputs, they still struggle to maintain 3D consistency across different views. 
In this paper, we present Consistent-1-to-3, which is a generative framework that significantly mitigates this issue.
Specifically, we decompose the NVS task into two stages: (i) transforming observed regions to a novel view, and (ii) hallucinating unseen regions. We design a scene representation transformer and view-conditioned diffusion model for performing these two stages respectively. Inside the models, to enforce 3D consistency, we propose to employ epipolar-guided attention to incorporate geometry constraints, and multi-view attention to better aggregate multi-view information. 
Finally, we design a hierarchy generation paradigm to generate long sequences of consistent views, allowing a full $360^{\circ}$ observation of the provided object image.
Qualitative and quantitative evaluation over multiple datasets demonstrates the effectiveness of the proposed mechanisms against state-of-the-art approaches. Our project page is at \url{https://jianglongye.com/consistent123/}.
\end{abstract}    
\section{Introduction}
\label{sec:intro}

\begin{table*}[htbp]
\centering
\setlength{\tabcolsep}{4pt}
\def\arraystretch{1}
\resizebox{0.85\linewidth}{!}{
\begin{tabular}{lccccc|ccccccc|ccc|ccccc|c}
\toprule
 & \multicolumn{5}{c}{Single instance}
 & \multicolumn{7}{c}{Re-projection}
 & \multicolumn{3}{c}{Latent}
 & \multicolumn{5}{c}{Diffusion}
 & Ours
 \\
 \cmidrule(lr){2-6}
 \cmidrule(lr){7-13}
 \cmidrule(lr){14-16}
 \cmidrule(lr){17-21}
 \cmidrule(lr){22-22}
 &
  \rot{NeRF~\cite{DBLP:journals/cacm/MildenhallSTBRN22}} &
  \rot{RegNeRF~\cite{DBLP:conf/cvpr/NiemeyerBMS0R22}} &
  \rot{VolSDF~\cite{DBLP:conf/nips/YarivGKL21}} &
  \rot{RealFusion~\cite{melas2023realfusion}} &
  \rot{3D Fuse~\cite{DBLP:journals/corr/abs-2303-07937}} &
  \rot{IBRNet~\cite{DBLP:conf/cvpr/WangWGSZBMSF21}} &
  \rot{PixelNeRF~\cite{DBLP:conf/cvpr/YuYTK21}} &
  \rot{NerFormer\cite{DBLP:conf/iccv/ReizensteinSHSL21}} &
  \rot{GPNR~\cite{DBLP:conf/eccv/SuhailESM22}} &
  \rot{VisionNeRF~\cite{Lin2022VisionTF}} &
  \rot{HumanVE~\cite{zhu2018view}} &
  \rot{PetNVS ~\cite{sinha2023common}} &
  \rot{LFN~\cite{DBLP:conf/nips/SitzmannRFTD21}} &
  \rot{SRT~\cite{DBLP:conf/cvpr/SajjadiMPBGRVLD22}} &
  \rot{OSRT~\cite{Sajjadi2022ObjectSR}} &
  \rot{3DiM~\cite{DBLP:conf/iclr/WatsonCMHT023}} &
  \rot{SparseFusion~\cite{zhou2023sparsefusion}} &
  \rot{NerfDiff~\cite{gu2023nerfdiff}} &
  \rot{GeNVS~\cite{chan2023genvs}} &
  \rot{Zero-1-to-3~\cite{DBLP:journals/corr/abs-2303-11328}} &
  \rot{Consistent-1-to-3} \\
\cmidrule{2-22}

{1) Single-view}                  & \n & \n & \n & \y & \y & \n & \y & \y & \n & \y & \y & \n & \n & \n & \n & \y & \n & \y & \y & \y & \y \\
{2) Sparse-views (2-4)}           & \n & \y & \y & \y & \y & \n & \y & \y & \n & \y & \y & \n & \n & \y & \y & \y & \y & \y & \y & \y & \y \\
{3) Multi-view consistent}        & \y & \y & \y & \y & \y & \y & \y & \y & \y & \y &  \y & \y & \y & \y & \y & \y & \y & \y & \y & \n & \y \\
{4) Generate unseen}              & \n & \n & \n & \y & \y & \n & \n & \n & \n & \n & \y & \y & \n & \n & \n & \y & \y & \y & \y & \y & \y \\
{5) Open-set generalization}      & \n & \n & \n & \y & \y & \n & \n & \n & \n & \n & \n & \n & \n & \n & \n & \n & \n & \n & \n & \y & \y \\
{6) Train-free for new instances} & \n & \n & \n & \n & \n & \y & \y & \y & \y & \y & \y & \y & \y & \y & \y & \y & \y & \y & \y & \y & \y \\
\bottomrule
\end{tabular}
}
\vspace{-3mm}
\caption{
    \textbf{Comparison with prior methods.}
    The rows indicate whether each method:
    1) works with a single input view,
    2) works with sparse (2-4) input views,
    3) generate consistent multiple views,
    4) hallucinates unseen regions,
    5) generalizes to open-set instances instead of limited categories,
    6) free of time-consuming training for new instances. Our proposed method stands out as the only approach that possesses all these advantages.
}
\label{tab:baselines}
\vspace{-3mm}
\end{table*}

Synthesizing high-quality and visually consistent novel views of real-world objects from a single input image has been a long-standing problem in computer vision. Its applications are widespread, ranging from content creation~\cite{DBLP:conf/icra/LinFBLRI22,DBLP:conf/iclr/PooleJBM23}, robotic manipulation and navigation~\cite{DBLP:conf/corl/MoreauPTSF21}, to 
AR/VR~\cite{DBLP:journals/tog/KopfMAQGCPFWYZH20}. 

This task is challenging as it involves difficulties from both reconstruction and generation, which require not only accurate geometry transformation of observed parts but also hallucination of unseen regions. 
In recent years, the field has seen significant progress with the advance of large-scale data-driven deep 2D models. For example, transformer models for single view reconstruction, \eg, DPT~\cite{ranftl2021vision},  GANs~\cite{goodfellow2020generative} and 2D diffusion models for image generation~\cite{DBLP:conf/nips/DhariwalN21}.  

To tackle such a challenge, one set of works requires a time-consuming iterative optimization to obtain novel view images through building an object-specific model
such as RealFusion~\cite{melas2023realfusion}, NeuralLift~\cite{Xu_2022_neuralLift} and SSDNerf~\cite{ranade2022ssdnerf}, where the input image needs to be involved in a multi-round training process. We note them as training-based methods.  Another set of works targets fast and efficient synthesis, which can complete the task with a single forward process of the given image, \eg, MCC~\cite{wu2023multiview}, Graf~\cite{schwarz2020graf} and Zero123~\cite{DBLP:journals/corr/abs-2303-11328}. We note them as training-free methods. In this paper, our method falls into the second category since more efficiency implies a wider range of applications. 
In principle, the effectiveness and generalization of training-free methods come from using a deep network to learn the prior information from a sufficiently large amount of 2D/3D data. 
For example, one may adopt large transformers~\cite{DBLP:conf/iccv/ReizensteinSHSL21, wu2023multiview} or diffusion-based models~\cite{chan2023genvs,DBLP:journals/corr/abs-2303-11328} for view synthesis.  

Among training-free methods, we are particularly interested in Zero123~\cite{DBLP:journals/corr/abs-2303-11328}, since it is among the recent models that can produce SoTA NVS quality, and generalize well to out-of-distribution images. We will discuss other models in related works in Sec.~\ref{sec:related-work}.
Specifically, Zero123 finetunes a pre-trained image-to-image diffusion model~\cite{sdvariation} on a large CAD dataset~\cite{deitke2023objaverse} with a novel camera embedding condition in the diffusion model. Although the quality is impressive, as shown in the top rows of Fig.~\ref{fig:teaser},  we found the results are still not geometrically consistent. 
While we conjecture this issue could be largely mitigated if given enough training data such as 10x bigger than objaverse-XL~\cite{deitke2023objaversexl} and adopted even larger pre-trained models such as Stable-Diffusion-XL~\cite{podell2023sdxl},  the current model for NVS is still resource-constrained, and the architecture design is essential for the quality of synthesized images. 

Therefore, we take a further step towards architecture optimization by designing a model that handles the challenges from reconstruction and generation separately, which significantly improves the geometric consistency, as shown in the bottom rows of Fig.~\ref{fig:teaser}.
To motivate our architecture design, we did an in-depth study of the Zero123 model, 
and concluded that the inconsistency comes from the inherent uncertainty of the diffusion-based model where a non-deterministic process is performed given a noisy input. 
Therefore, it is difficult to maintain the 3D consistency across independently sampled views. 
To tackle this issue, we introduce a two-stage model, as shown in Fig.~\ref{fig:method}. In the first stage, we set up a deterministic Scene Representation Transformer (SRT)~\cite{DBLP:conf/cvpr/SajjadiMPBGRVLD22} which quickly transforms the observed image into novel views that roughly represent the object's shape and appearance. In the second stage, we design a view-conditioned diffusion model, which is conditioned not only on the camera embedding but also the outputs from SRT, to produce clear and detailed images for unseen regions.
To enforce 3D consistency, we additionally employ epipolar-guided attention~\cite{tseng2023consistent} and multi-view attention~\cite{shi2023MVDream} to better aggregate multi-view information.  Finally,  a hierarchy generation paradigm is proposed to generate long sequences of consistent 3D views, allowing a full experience of $360^{\circ}$ observation of the provided object image. We name this architecture as Consistent 1-to-3 and experiment with it over several popular datasets~\cite{deitke2023objaverse}. In all cases, our results outperform other SoTA methods in terms of both quality and consistency,  demonstrating the effectiveness of the proposed method. Finally, Consistent 1-to-3 can further improve the NVS performance when using few-shot images as input.

In summary, our paper contributes a novel geometric-aware deep model, namely Consistent 1-to-3,  for efficient novel view synthesis.  
With extensive experiments on multiple datasets, it produces state-of-the-art synthesis quality and geometric consistency across various camera viewpoints. 

\section{Related Work}
\label{sec:related-work}
\vspace{-1mm}

In this section, we briefly review the literature in novel-view synthesis and group related methods based on their attributes, shown
in Tab.~\ref{tab:baselines}. 
In the following, we will walk through these works on NVS to show the uniqueness of Consistent 1-to-3.   
We recognize the breadth and diversity within NVS, and only include
representative works due to limited space. 
While we acknowledge the existence of other significant contributions, citing them all is beyond our scope.

\noindent
\textbf{Single Instance Training-based Reconstruction.}  As discussed in Sec.~\ref{sec:intro},
this thread of works introduces a multi-round optimization strategy to obtain an instance-specific model for view synthesis.  NeRF~\cite{mildenhall2021nerf} first proposes a neural network that optimizes around 100 images to produce realistic synthesized images.   VolSDF~\cite{DBLP:conf/nips/YarivGKL21} can train a NeRF model with tens of images by incorporating a 3D volume representation.  Later works try to further reduce the number of required images, while maintaining the original rendering quality.  RegNeRF~\cite{DBLP:conf/cvpr/NiemeyerBMS0R22} takes 3 images and adopts a geometry loss from depth patches.  SinNerf~\cite{Xu_2022_SinNeRF}, RealFusion~\cite{melas2023realfusion} and NeuralLift~\cite{Xu_2022_neuralLift} further 
reduce the required images to 
a single one by combining either depth map or Score Distillation Sampling~\cite{DBLP:conf/iclr/PooleJBM23} during the nerf model training.  
However, all these strategies need time-consuming optimization, \eg, more than 30 mins for a single model for RealFusion, which hinders a wide range of interactive applications.

\noindent
\textbf{Re-projection methods.} This thread of works considers adopting an explicit geometry representation, then doing a 3D re-projection for novel view synthesis.
IBRNet~\cite{wang2021ibrnet} and GPNR~\cite{DBLP:conf/eccv/SuhailESM22} adopts image-based view synthesis using a sparse set of nearby views with pixel or patch-based embedding. PixelNeRF~\cite{DBLP:conf/cvpr/YuYTK21} trained a generalizable Nerf model with 13 object categories from ShapeNet~\cite{Chang2015ShapeNetAI} while not showing its ability to learn from large amount of instances in open-set images such as
Objaverse~\cite{deitke2023objaverse}.  NerFormer~\cite{DBLP:conf/iccv/ReizensteinSHSL21} further enlarges the 3D datasets to 50 categories with 18,619 instances and proposes a transformer for view synthesis from a small number of images. Later, MCC~\cite{wu2023multiview} introduce a volume representation yielding better performance when a single image is given, but the rendered images lack details. In addition to category-agnostic approaches, some works adopt category-specific mesh models as prior to improve the results such as Human View Extrapolation (HumanVE)~\cite{zhu2018view} or Pet Novel View  Synthesis (PetNVS)~\cite{sinha2023common}. These models can hardly generalize to open-set instances. 

\noindent
\textbf{Latent methods.} This thread of work generalizes the explicit pixel representation to latent ones and learns MLP to directly decode the latent code to novel-view rather than performing ray aggregation from NeRF, yielding much faster rendering speed. This is first proposed in LFN~\cite{DBLP:conf/nips/SitzmannRFTD21}. Later, SRT~\cite{DBLP:conf/cvpr/SajjadiMPBGRVLD22} further reduced the necessary observed images to a few shots. However, these methods have not shown their generalization ability across open-set categories. 

\noindent
\textbf{Diffusion Models for NVS.} 
This thread of work extends existing image diffusion models to conditional diffusion models where input views and target camera parameters are used as the conditions for generating a target view. Watson \etal~\cite{DBLP:conf/iclr/WatsonCMHT023} first proposed to use diffusion models for such a task by training a diffusion model on the SRN ShapeNet dataset~\cite{sitzmann2019scene}. SparseFusion~\cite{zhou2023sparsefusion} combines stable diffusion model~\cite{DBLP:conf/cvpr/RombachBLEO22} with epipolar feature transformer for a view-conditioned diffusion model in the latent space. GeNVS~\cite{chan2023genvs} proposed to improve the view consistency of the diffusion model by re-projecting the latent feature of input view before diffusion denoising. However, all of these works are limited to their training data and have not been proven to generalize to arbitrary image input. Zero-1-to-3~\cite{DBLP:journals/corr/abs-2303-11328} utilizes pre-trained stable diffusion model~\cite{sdvariation} and fine-tunes it on a large 3D render dataset. It is found that such a fine-tuned model is able to generate high-quality novel view output with relatively good generalizability compared to prior works. The follow-up work One-2-3-45~\cite{Liu2023One2345AS} takes advantage of the generalization capabilities of Zero-1-to-3 and employs a feed-forward network to obtain a mesh representation. More recently, MVDream~\cite{shi2023MVDream} proposes a similar multi-view diffusion mechanism to generate geometrically consistent multi-view images from a given text prompt. 

\noindent
\textbf{3D Diffusion Models.}
Utilizing categorical 3D datasets, numerous works have leveraged diffusion models directly on 3D shapes. Zhou \etal~\cite{zhou20213d} and Zeng \etal~\cite{vahdat2022lion} employ diffusion models for 3D point clouds. $PC^2$~\cite{melas2023pc2} learns a point cloud generation model conditioned on a single image.
Diffusion models based on various 3D representations, such as meshes~\cite{liu2023meshdiffusion}, volumes~\cite{cheng2023sdfusion}, and neural implicit representations~\cite{gupta20233dgen, chou2022diffusionsdf}, have also been explored. 
We have also seen rapid progress on the more challenging open-set (or universal) 3D generation~\cite{yu2023pushing} thanks to the release of Objaverse~\cite{deitke2023objaverse}, a large-scale 3D model repository. 
Researchers from OpenAI also proposed Point-e~\cite{nichol2022point}, and Shap-e~\cite{jun2023shap} for open-set 3D generation conditioned on texts or images using internal 3D datasets. 

\section{Method}
\label{sec:method}

\begin{figure*}[t]
  \centering
  \includegraphics[width=1.0\textwidth]{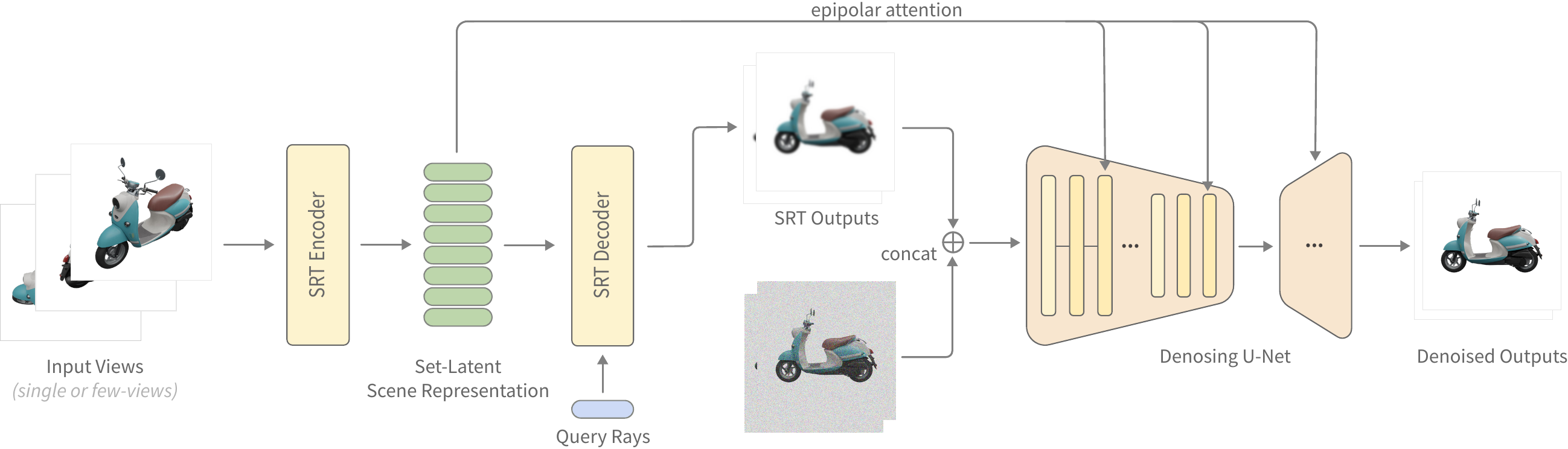}
  \vspace{-0.1in}
  \caption{\textbf{Pipeline of Consistent-1-to-3}. Given a single or a sparse set of input images, the encoder within Scene Representation Transformer (SRT) translates the image(s) into latent scene representations, effectively capturing implicit 3D information. The image rendering unfolds in two stages. The initial stage produces a rough yet geometry-grounded output by cross-attending the queried pixels to the latent scene representation. Subsequently, these intermediate outputs are taken as input by the view-conditioned diffusion model, resulting in visually appealing images that exhibit consistency among the input and generated images from different viewpoints.}
  \vspace{-0.1in}
  \label{fig:method}
\end{figure*}

Given a single observation of an object, denoted as $\boldsymbol{x}^{\mathrm{source}}$, with known camera parameters $\boldsymbol{\pi}^{\mathrm{source}}$, our goal is to synthesize a set of multi-view consistent images using target camera parameters $\boldsymbol{\pi}^{\mathrm{target}}$.
As illustrated in~\cref{fig:method}, our pipeline consists of two key components: 1) A Scene Representation Transformer that learns a latent 3D representation given a single view; and 2) a view-conditioned diffusion model to capture the generative aspects of novel view synthesis.
We first detail the pipeline in~\cref{subsec:geo-nvs}, and introduce several essential techniques to further enhance multi-view consistency in ~\cref{subsec:mv-consistency}. 
Lastly, we extend the proposed Novel View Synthesis pipeline to facilitate 3D reconstruction (~\cref{subsec:3d-recon}). 

\subsection{Geometry-guided Novel View Synthesis}
\label{subsec:geo-nvs}
We decompose the NVS task into two sub-tasks: (i) photometric warping, which is guided by epipolar constraints to ensure accurate alignment of observed regions, and (ii) hallucination of unseen regions based on observed regions. 
To accomplish these sub-tasks, we utilize a scene representation transformer for photometric warping and a view-conditioned diffusion model for the generation of unseen regions. 

\noindent
\textbf{Scene Representation Transformer (SRT)}. SRT~\cite{DBLP:conf/cvpr/SajjadiMPBGRVLD22} builds upon a transformer encoder-decoder architecture to learn an implicit 3D representation given a set of posed images $\left(\boldsymbol{x}^{\mathrm{source}}, \boldsymbol{\pi}^{\mathrm{source}}\right)$:
\begin{align}
    \boldsymbol{z} &= f_{e} \circ f_{c} \left(\boldsymbol{x}^{\mathrm{source}}, \boldsymbol{\pi}^{\mathrm{source}}\right),
\end{align}
where a CNN backbone $f_{c}$ extracts patch-level features from posed input images and feeds them as tokens to the transformer encoder $f_{e}$. The encoder converts input tokens to a set-latent scene representation $\boldsymbol{z}$ via self-attention.

To render a novel image of the scene represented by $\boldsymbol{z}$, the decoder of SRT queries the pixel color via cross attention between the ray corresponding to the pixel $\boldsymbol{r} \equiv (\boldsymbol{o}, \boldsymbol{d})$ with $\boldsymbol{o}$ and $\boldsymbol{d}$ being the origin and normalized direction of the ray and the set-latent scene representation $\boldsymbol{z}$:
\begin{align}
    \hat{C}(\boldsymbol{r}) &= f_{d} (\boldsymbol{r}, \boldsymbol{z}),
\end{align}
where $f_{d}$ is the decoder of SRT.

Several changes are necessary to integrate SRT into our pipeline. 
Specifically, we replace all standard attention layers with epipolar attention, which allows the model to incorporate important geometry biases (details explained below). 
Furthermore, we train the SRT in the image latent space, so that it not only reduces computation costs but also ensures seamless compatibility with the latent diffusion model.

Our SRT is trained by minimizing a pixel-level reconstruction loss:
\begin{align}
  \mathcal{L}_{\mathrm{recon}} =\sum_{\mathbf{r} \in \mathcal{R}}\left\|C(\mathbf{r})-\hat{C}(\mathbf{r})\right\|_2^2,
\end{align}
where $C(\mathbf{r})$ is the ground truth color of the ray and $\mathcal{R}$ is the set of rays sampled from target views.

\noindent
\textbf{View-conditioned Diffusion}. 
While SRT is able to learn geometry correspondence via the attention mechanism, the pixel-level reconstruction loss leads to averaging out fine-grained details in the images, thus often resulting in blurred predictions. 
To capture the probabilistic nature of novel view synthesis (for the unseen parts), we employ a view-conditioned diffusion model to estimate the conditional distribution of the target view given the source view and the relative camera pose: $p \left(\boldsymbol{x}^{\mathrm{target}} \mid \boldsymbol{\pi}^{\mathrm{target}}, \boldsymbol{x}^{\mathrm{source}}, \boldsymbol{\pi}^{\mathrm{source}} \right)$. 
Note that although we follow LDM~\cite{DBLP:conf/cvpr/RombachBLEO22} to perform the diffusion process in the image latent space, we still denote the image in latent space with $\boldsymbol{x}$ for simplicity. 

To achieve conditional generation, our denoising network $\boldsymbol{\epsilon}_\theta$ operates under two conditions. 
Firstly, the SRT predicts a $32 \times 32$ latent image $\tilde{\boldsymbol{x}}^{\mathrm{target}}$ based on the target view $\boldsymbol{\pi}^{\mathrm{target}}$. 
This predicted latent image is concatenated with the noisy image $\boldsymbol{y}$ and fed into the denoising network. 
Secondly, the denoising network is also conditioned on the scene representation $\boldsymbol{z}$ through multiple cross-attention layers.
By incorporating these two conditions, the denoising network can effectively leverage both the target view information and the scene representation to produce sharp and realistic images that are also consistent with the input images.
Concretely, the predicted images $\hat{\boldsymbol{\epsilon}_t}$ are given by:
\begin{align}
   \hat{\boldsymbol{\epsilon}_t} &= \boldsymbol{\epsilon}_\theta (\boldsymbol{y}, \tilde{\boldsymbol{x}}^{\mathrm{target}}, \boldsymbol{z}, t),
\end{align}
where $t$ is the timestep.

The network is trained by optimizing a simplified variational lower bound:
\begin{align}
  \mathcal{L}_{\mathrm{diffusion}}=\mathbb{E}\left[\left\|
      \boldsymbol{\epsilon}_t - \boldsymbol{\epsilon}_\theta (\boldsymbol{y}, \tilde{\boldsymbol{x}}^{\mathrm{target}}, \boldsymbol{z}, t)
  \right\|^2\right]
\end{align}

\begin{figure*}[t]
  \centering
  \includegraphics[width=1.0\textwidth]{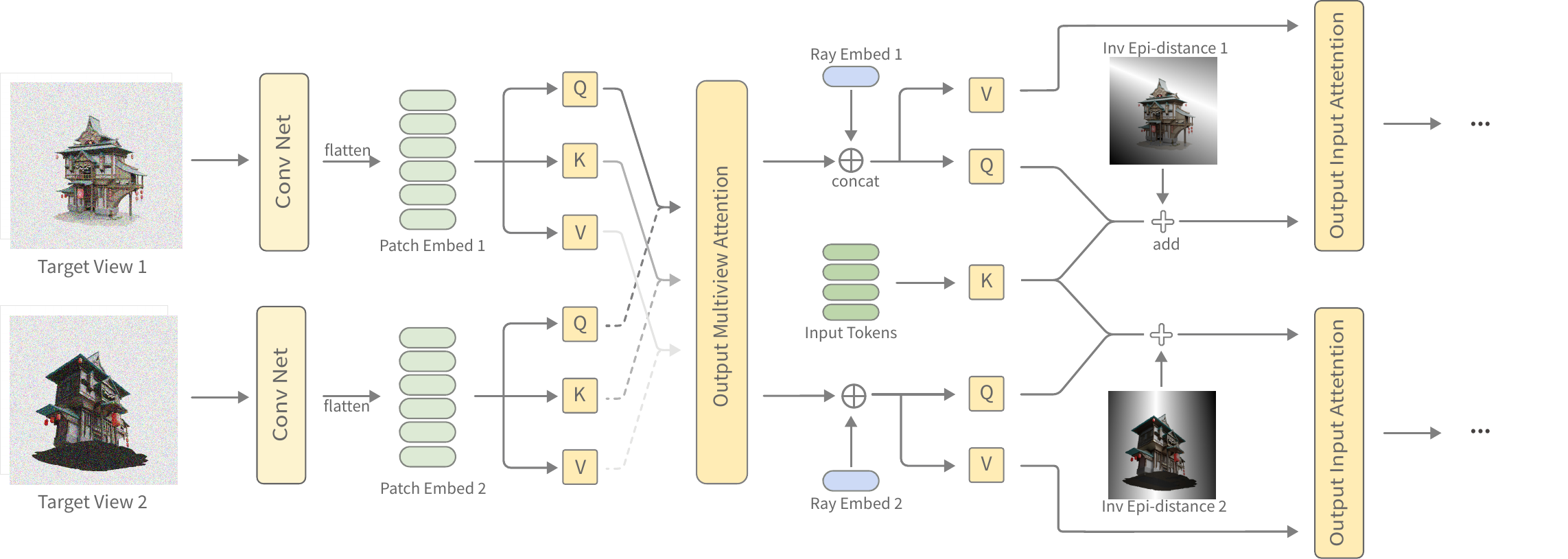}
  \vspace{-0.15in}
  \caption{\textbf{Details of UNet Block}. After image features are extracted by CNN, they are aggregated via the multi-view attention mechanism in the transformer. The output images are conditioned on 1) the multi-view feature; 2) the queried ray; and 3) the affinity matrix reweighted by the epipolar attention. }
  \label{fig:unet}
  \vspace{-3mm}
\end{figure*}

\noindent
\textbf{Epipolar Attention}. 
While previous works on the view-conditioned diffusion model~\cite{DBLP:journals/corr/abs-2303-11328} employ full attention in their model, we propose to exploit the valuable information between images through relative camera poses using epipolar constraints. To this end, we replace all cross-attention layers in our SRT and diffusion model between source and target views with our novel epipolar attention mechanism. 
In our implementation of epipolar attention, we reweigh the affinity matrix $A_{i, j}$ between the $i^{th}$ and $j^{th}$ view by taking the epipolar distance into account.
For each pixel in view $i$, we compute the epipolar line in view $j$ and determine the epipolar distance for all pixels in view $j$. Subsequently, we convert the inverse epipolar distance into a weight map $K_{i, j}$. The affinity matrix is then reweighted using $A_{i, j}^\prime = A_{i, j} + K_{i, j}$.

\subsection{Improving Multi-view Consistency}
\label{subsec:mv-consistency}
To generate smooth Novel View Synthesis (NVS) images, it is essential to ensure not only the consistency between the generated views and input view but also the consistency among the generated views themselves. 
While our novel pipeline mentioned earlier primarily emphasizes the consistency between generated and input views, we introduce additional beneficial techniques to enhance the consistency among the generated views. 
These techniques are designed to further improve the quality and coherency of the synthesized views, providing more visually pleasing and accurate images.

\noindent
\textbf{Multi-view Attention}. 
In contrast to previous methods, such as Zero123~\cite{DBLP:journals/corr/abs-2303-11328}, which generate multiple images of a givens object instance sequentially, we have observed that this sequential approach can introduce inconsistencies between generated views due to randomness in the multiple backward passes of the diffusion model.
To address this issue, we propose an alternative approach to learn
the joint distribution of multi-view images. 
This means predicting several novel views simultaneously given the source image $\boldsymbol{x}^{\mathrm{source}}$. 
To achieve this, as demonstrated in ~\cref{fig:unet}, we modify the UNet architecture by feeding a batch of noisy multi-view images during the forward pass and extending the self-attention block to work with multi-view images such that information of different viewpoints is aggregated to produce consistent rendered images. 

\noindent
\textbf{Hierarchy Generation}. 
To generate novel views that are far from the input viewpoint, one naive approach is to generate the target view directly. 
Nonetheless, this approach is less effective due to self-occlusion that caused unreliable epipolar matching.
Instead, we present a hierarchical approach for generating extensive sequences of novel views. 
Specifically, we generate new images conditioned on not only the input view but also previously generated views.

\subsection{3D Reconstruction }
\label{subsec:3d-recon}
While consistent NVS is great for 3D assets preview and many other applications, many scenarios within AR/VR still require accurate 3D
objects for occlusion/lighting reasoning that is attributed to realistic and immersive experience.

Following Zero123~\cite{DBLP:journals/corr/abs-2303-11328}, we adopt an optimization-based method to generate 3D assets using Score Distillation Sampling (SDS)~\cite{DBLP:conf/iclr/PooleJBM23}.
The fundamental concept behind SDS optimization involves updating the 3D representation, often in the form of a NeRF model, by maximizing the likelihood of rendered views using a trained 2D diffusion model.
When applying our model to the SDS optimization process, we sample a set of random camera poses instead of a single one to accommodate the multi-view attention mentioned earlier.

\begin{figure*}[htbp]
  \centering
  \includegraphics[width=1.0\textwidth]{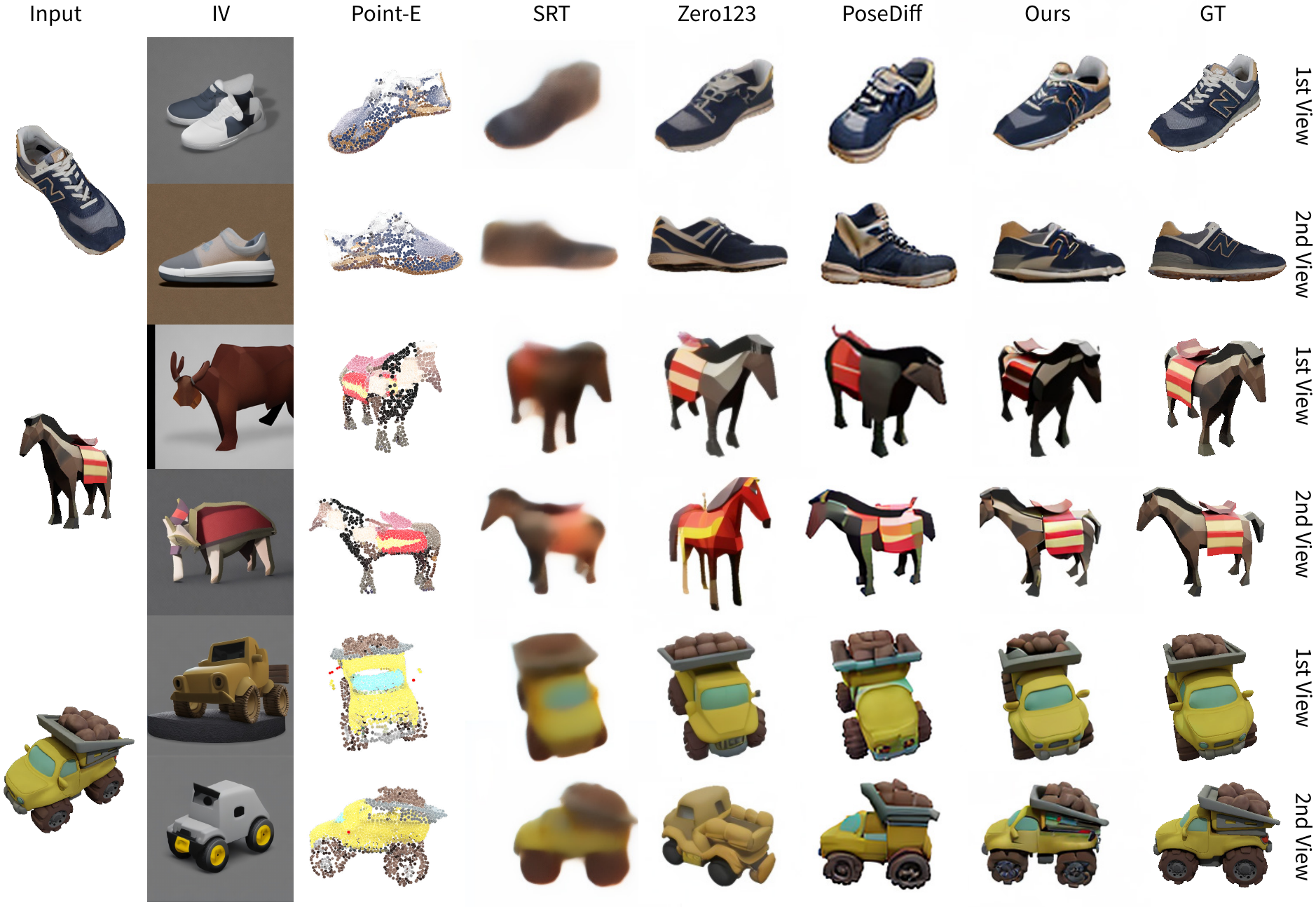}
  \caption{\textbf{NVS visualization.} We visualize the NVS results across a wide range of methods on the Objaverse dataset. Our approach outperforms the competitive baselines by producing both realistic and consistent images.}
  \label{fig:nv}
\end{figure*}

\begin{figure*}[htbp]
  \centering
  \includegraphics[width=1.0\textwidth]{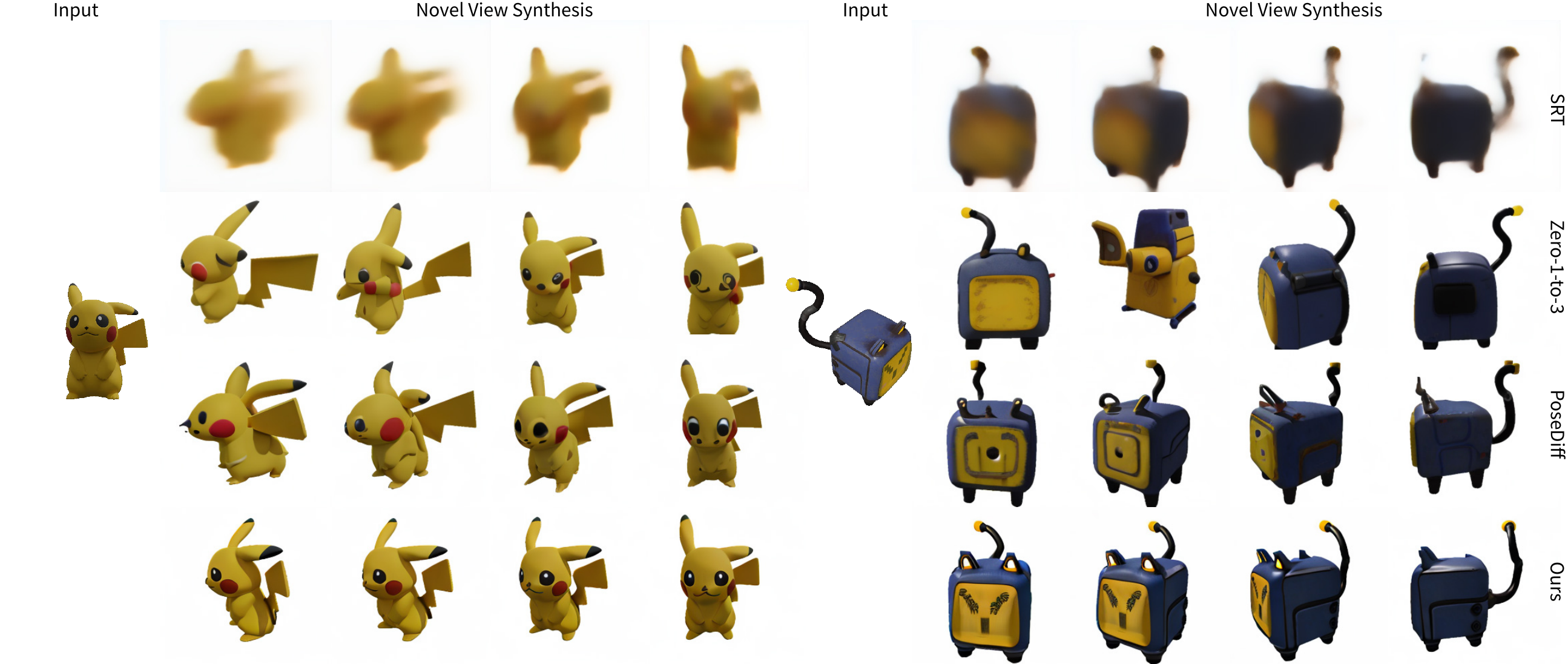}
  \caption{\textbf{Qualitative results on NVS Consistency}. While SRT (the first row) can generate views that are geometrically consistent with the input image, the rendered images are blurry. Previous diffusion-based approaches (the second and third row) are able to generate sharp and realistic-looking images, but the diffusion process leads to inconsistency. Our geometry-guided approach gets the best of both worlds, generating not only plausible but also consistent images. }
  \label{fig:consistency}
\end{figure*}

\section{Experiments}
\label{sec:exp}

\subsection{Experimental Setting}

\noindent
\textbf{Dataset}.
We fine-tune our diffusion model on the recently released Objaverse~\cite{deitke2023objaverse} dataset, which is a large-scale CAD dataset containing 800K high-quality objects. We directly employ the processed rendering data from Zero123, which provides 12 random views for each object. For evaluation, we use the test split of Objaverse provided by Zero123. In addition, to test our model's performance on the out-of-distribution data, we also evaluate on the Google Scanned Objects~\cite{DBLP:conf/icra/DownsFKKHRMV22} dataset, which contains high-quality scanned household items. Images in all datasets are resized to $256 \times 256$. The background is set to white.

\noindent
\textbf{Metrics}.
Following previous works~\cite{DBLP:journals/corr/abs-2303-11328}, we use Peak signal-to-noise ratio (PSNR), Structural Similarity Index (SSIM),  and Learned Perceptual Image Patch Similarity (LPIPS) to measure the similarity between rendered images and ground truth images. Moreover, we use the flow warping error $E_{\mathrm{warp}}$~\cite{DBLP:conf/eccv/LaiHWSYY18} to quantify the consistency across different views. Specifically, we generate a 32-frame video along a pre-defined circle of camera trajectory for each object and then use the RAFT model~\cite{DBLP:conf/eccv/TeedD20} to compute the optical flow between two consecutive generated frames. The 
error $E_{\mathrm{warp}}$ is defined as
$$
E_{\mathrm{warp}} = \sum_i \left\|\boldsymbol{x}^i-\hat{\boldsymbol{x}}^{i-1}\right\|_1,
$$
where $\hat{\boldsymbol{x}}^{i-1}$ is warped from the generated frame $\boldsymbol{x}^{i-1}$ using the optical flow.

\noindent
\textbf{Baselines}. We mainly evaluate our approach against methods that can generalize to open-set categories and accept single-view RGB images as inputs. In particular, we measure our performance against Zero123~\cite{DBLP:journals/corr/abs-2303-11328}, the first work that fine-tune the view-conditioned diffusion model on Objaverse, showcasing impressive zero-shot ability. Following Zero123, we also report results of Image Variation~\cite{sdvariation}, a Stable Diffusion model conditioned on images, which is also the pre-trained model we fine-tuned from. Besides, we adapt Pose-Guided Diffusion~\cite{tseng2023consistent}, a recent diffusion model for generating consistent videos using an epipolar attention mechanism similar to ours. Since the code wasn't public, we re-implement and re-train it on Objaverse. We also add a camera embedding the same as ours to the model for more precise view control.

\noindent
\textbf{Implementation Details}. Both SRT encoder and decoder are adapted from ViT-Base~\cite{DBLP:conf/iclr/DosovitskiyB0WZ21}. In encoder, the attention layers has 12 heads and 4 layers, the decoder's attention layers consist of 6 heads and 3 layers. The ray for each image patch
is parameterized in Plucker coordinates~\cite{DBLP:conf/nips/SitzmannRFTD21} and the dimension of ray embedding is 120. The MLPs in SRT have 1 hidden layer and GELU activations~\cite{hendrycks2016gaussian}. The VAE modules in diffusion models are freezed during training. The downsample scale of VAE is 8, which results in $32\times32$ latent images. The basic UNet module is borrowed from Stable Diffusion. We use AdamW~\cite{DBLP:conf/iclr/LoshchilovH19} with a learning rate of $5 \times 10^{-5}$ for training. The model is fine-tuned with a batch size of 1024 on 32$\times$A100-80GB for 7 days.

\begin{table}[t]
\centering
\small
\setlength{\tabcolsep}{5pt}
\def\arraystretch{1}
\resizebox{\linewidth}{!}{
\scriptsize
\begin{tabular}{lcccc}
	\toprule
	& $\uparrow$\,PSNR
	& $\uparrow$\,SSIM
	& $\downarrow$\,LPIPS
	& $\downarrow$\,$E_{\mathrm{warp}}$ \\
    \cmidrule(lr){2-5} 
	{ImageVariation~\cite{sdvariation}}
        & 6.42 & 0.506 & 0.578 & 22.4
	\\
	{Zero-1-to-3~\cite{DBLP:journals/corr/abs-2303-11328}}
	& 18.13 & 0.828 & 0.157 & 14.2 \\
	{Pose-Diffusion$^{*}$~\cite{tseng2023consistent}}
	& \sbest 18.92 & \sbest 0.841 & \sbest 0.141 & \sbest 7.0 \\
	{Ours} 
	& \best\bf{20.72} & \best\bf{0.877} & \best\bf{0.112} & \best\bf{3.7} \\
	\bottomrule
\end{tabular}
}
\vspace{-2mm}
\caption{
\textbf{Quantitative results on Objaverse.} We evaluate our method on the test split of Objaverse and demonstrate significant improvements across all metrics. Pose-Diffusion$^{*}$ is our re-implemented version.
}
\label{tab:nvs-objaverse}
\end{table}

\begin{table}[t]
\centering
\small
\setlength{\tabcolsep}{5pt}
\def\arraystretch{1}
\resizebox{\linewidth}{!}{
\scriptsize
\begin{tabular}{lcccc}
	\toprule
	& $\uparrow$\,PSNR
	& $\uparrow$\,SSIM
	& $\downarrow$\,LPIPS
	& $\downarrow$\,$E_{\mathrm{warp}}$ \\
    \cmidrule(lr){2-5} 
	{ImageVariation~\cite{sdvariation}}
        & 5.93 & 0.538 & 0.529 & 19.9
	\\
	{Zero-1-to-3~\cite{DBLP:journals/corr/abs-2303-11328}}
	& 16.72 & 0.721 & 0.212 & 11.7  \\
	{Pose-Diffusion$^{*}$~\cite{tseng2023consistent}}
	& \sbest 17.52 & \sbest 0.791  & \sbest 0.174 & \sbest 5.2  \\
	{Ours} 
	& \best\bf{19.46} & \best\bf{0.858} & \best\bf{0.146} & \best\bf{3.3} \\
	\bottomrule
\end{tabular}
}
\vspace{-2mm}
\caption{
\textbf{Quantitative results on Google Scanned Objects.}
Evaluation of novel-view synthesis on the out-of-distribution GSO dataset.
Ours is best across datasets and metrics.
}
\label{tab:nvs-gso}
\end{table}

\subsection{Novel View Synthesis}

We present quantitative results for NVS on both in-distribution and out-of-distribution datasets in Tab.~\ref{tab:nvs-objaverse} and ~\ref{tab:nvs-gso}. Our proposed mechanisms significantly improve the performance of all metrics. Fig.~\ref{fig:nv} shows a comparison of NVS results on Objaverse with our method and all baselines. The SRT~\cite{DBLP:conf/cvpr/SajjadiMPBGRVLD22} results are
obtained by training the SRT part of our model with only the reconstruction loss. On one side, Point-E and SRT fail to produce realistic images but remain consistent with the input; on the other side, diffusion-based baselines (Zero123 and PoseDiff) can generate realistic images but do not align well with the ground truth. Our approach is the only one that achieves both fidelity and consistency.

Fig.~\ref{fig:consistency} provides the qualitative results for the NVS consistency. We render a sequence of images using the same camera trajectory for each approach. While previous diffusion models (Zero123 and PoseDiff) are able to generate realistic novel views, they often suffer from inconsistent issues across the independently sampled views. By incorporating epipolar constraint and modeling the joint distribution of multi-views, our method significantly alleviates this issue.

\begin{figure}
  \centering
  \includegraphics[width=1.0\linewidth]{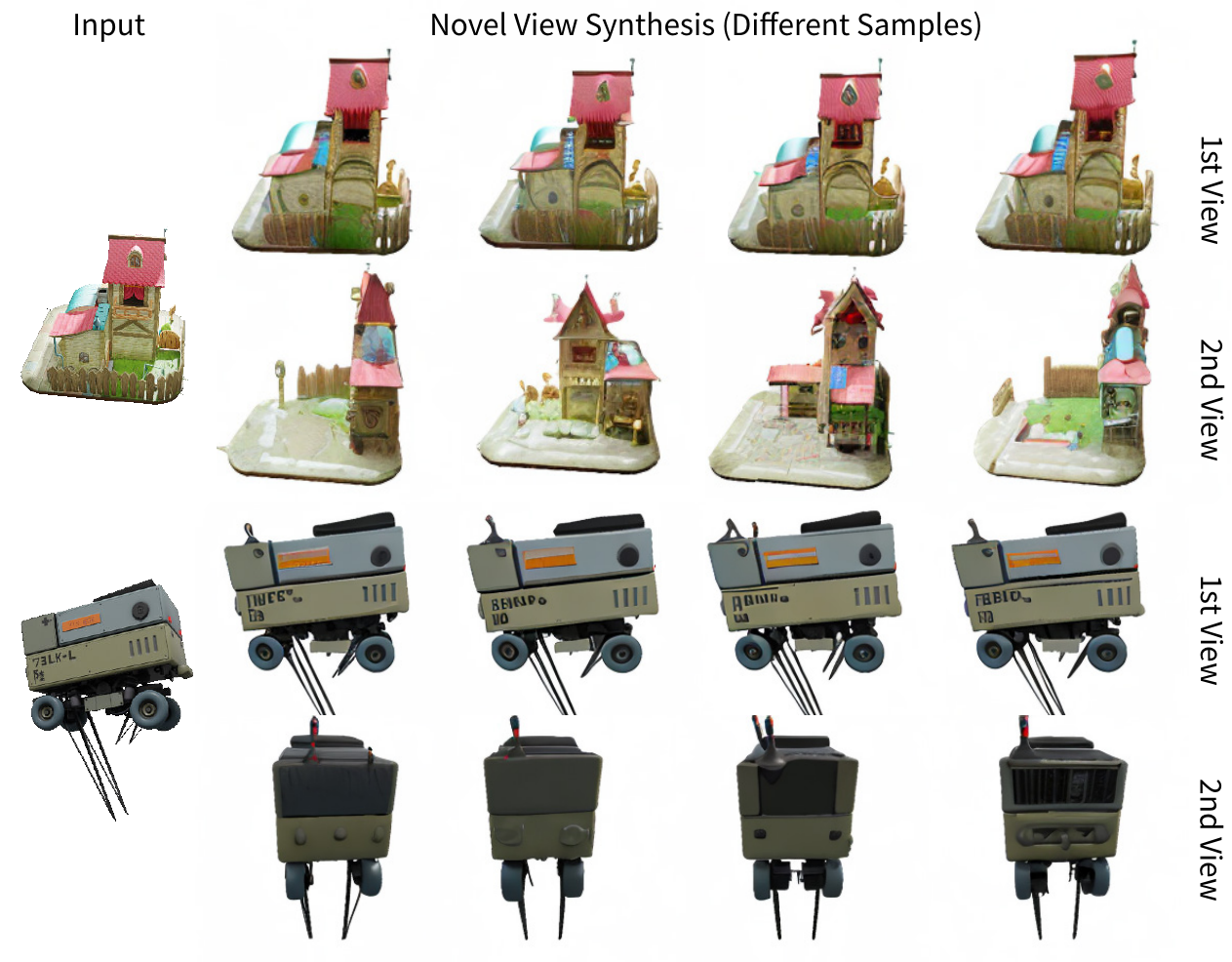}
  \caption{\textbf{Diversity of NVS results}. Nearby views (the 1st view) maintain consistency with the input image, while distant views (the 2nd view) generate a variety of expressive yet realistic images. Note the diverse patterns on the back of the trailer.}
  \label{fig:diversity}
\end{figure}

By learning the data distribution, diffusion approaches, by nature, can generate more diverse samples. But when dealing with NVS tasks, observed regions in target views must reflect the geometric transformation rather than being randomly generated. Our design is guided by this motivation, and Fig.~\ref{fig:diversity} shows the effectiveness of our design. The first view of each object is close to the input view, and all generated images align with the input. But for the unseen regions shown in the second view, the variation of generated images is much larger the first view.

\begin{figure}
  \centering
  \includegraphics[width=1.0\linewidth]{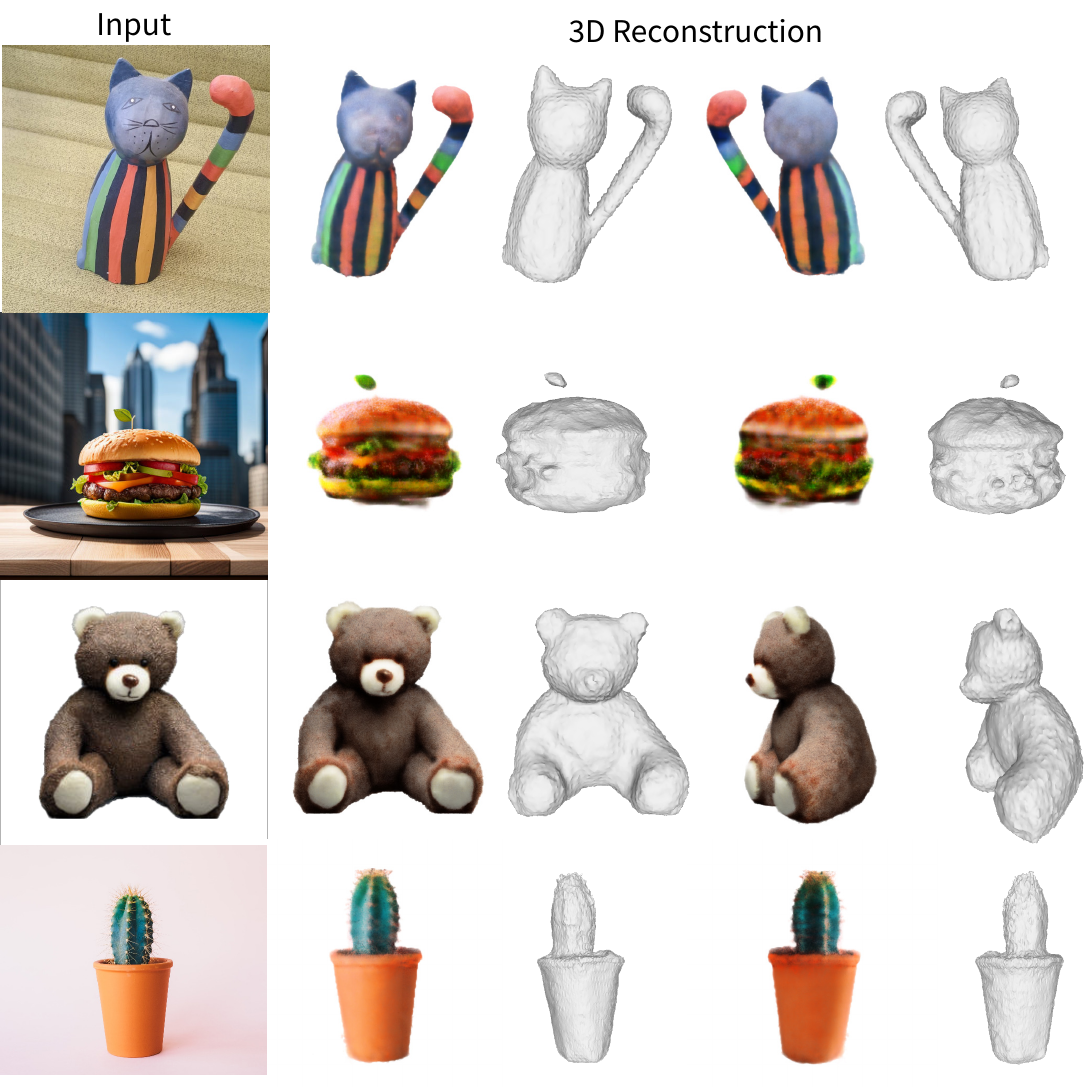}
  \caption{\textbf{3D reconstruction.} By combining the view-conditioned diffusion model and NeRF-style reconstruction techniques, our method is able to perform high-fidelity 3D reconstruction for any single image. }
  \label{fig:recon}
\end{figure}

\subsection{3D Reconstruction}

Fig.~\ref{fig:recon} qualitatively shows the 3D reconstruction results of our method. For each instance, we optimize a NeRF via the SDS loss, the colored images in the figure are all NeRF renderings. After reconstructing NeRFs, we further utilize DiffRast~\cite{Laine2020diffrast} to extract meshes from NeRFs. In this way, we leverage the multi-view prior from our diffusion models and combine them with the advanced NeRF-style reconstruction.

\begin{table}[t]
\centering
\small
\setlength{\tabcolsep}{5pt}
\def\arraystretch{1}
\resizebox{\linewidth}{!}{
\scriptsize
\begin{tabular}{lcccc}
	\toprule
	& $\uparrow$\,PSNR
	& $\uparrow$\,SSIM
	& $\downarrow$\,LPIPS
	& $\downarrow$\,$E_{\mathrm{warp}}$ \\
    \cmidrule(lr){2-5} 
        {w.o. epi attention} 
        & 19.34 & 0.847 & 0.148 & \sbest 4.3 \\
        {w.o. in-out attention} 
        & 18.79 & 0.823 & 0.156 & 4.6 \\
        {w.o. mv-out attention} 
        & \sbest 20.84 &  \sbest 0.879 &  \sbest 0.120 & 7.0 \\
        {w.o. hierarchy generation}
        & \best\bf{21.03} & \best\bf{0.883} & \best\bf{0.107} & 8.8 \\
	{full model} 
	& 20.72 & 0.877 & 0.112 & \best\bf{3.7} \\
	\bottomrule
\end{tabular}
}
\vspace{-2mm}
\caption{
\textbf{Ablation study on Objaverse}. Each of our design choices contributes to fidelity or consistency. Refer to the text for the detailed discussion.
}
\label{tab:ablation}
\end{table}

\subsection{Ablation Study}

We ablate each design choice in our method and present quantitative results in Tab.~\ref{tab:ablation}. The epipolar line guidance (1st row) and the attention between inputs and outputs (2nd row) enforce the model to learn the better geometric transformation of observed regions in source views, and thus they are essential to the similarity metrics (PSNR, SSIM, and LPIPS). The attention between multiple outputs (3rd row) and the hierarchy generation (4th row) contributes a lot to the consistency across different views. However, it should be noted that these two mechanisms will slightly hurt the reconstruction performance, and our full model is a trade-off between fidelity and consistency. 
 \section{Conclusion}
\label{sec:conclusion}
In this paper, we proposed Consistent 1-to-3, a geometry-aware deep model that produces high quality and 3D consistent novel view synthesis from a single image or few-shot images.  
Consistent 1-to-3 is a two-stage model that first uses a transformer to generate a blurry while geometry correct image, and then uses a diffusion model to finish the details. We demonstrate by following such a generation progression, the generated results are significantly improved. 
We hope our work could motivate future studies on incorporating better geometry constraints and representations inside the model, which could significantly improve its practical usage.

{
    \small
    \bibliographystyle{ieeenat_fullname}
    \bibliography{main}
}
\clearpage
\setcounter{page}{1}
\maketitlesupplementary

\section{Overview}

In this supplementary material, we first provide the technical details of the proposed method to complement the paper. Second, we describe the experiment details, including the metric computation and reproducing baselines. Finally, we present additional qualitative results of baselines and our approach. Please see our \textbf{video} for an overview of our work and more visualizations.

\section{Technical Details}
\label{sec:technical-details}

\begin{figure}[h!]
  \centering
  \vspace{-5 pt}
  \includegraphics[width=0.9\linewidth]{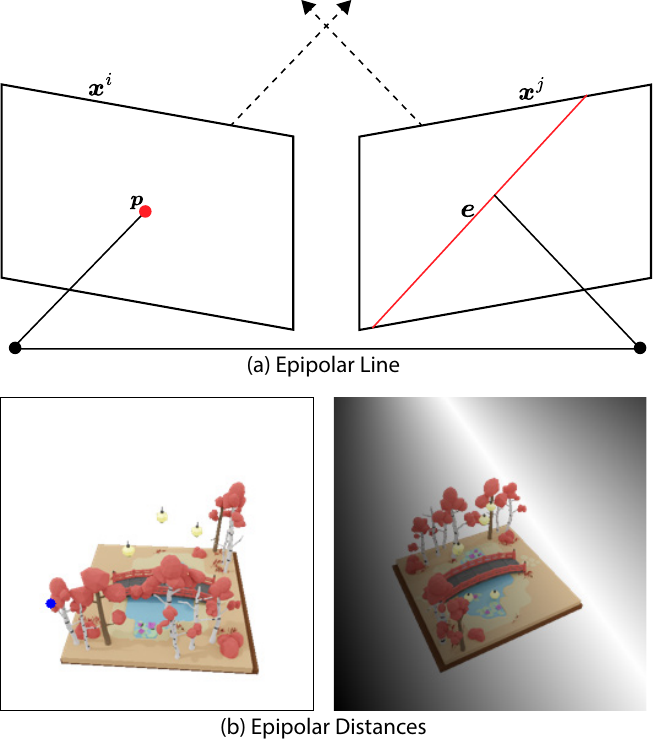}
  \vspace{-5 pt}
  \caption{By combining the view-conditioned diffusion model and NeRF-style reconstruction techniques, our method is able to preform high-fidelity 3D reconstruction for any single images. }
  \label{fig:epipolar}
  \vspace{-10 pt}
\end{figure}

\textbf{Epipolar Attention}. The epipolar line computation process is visualized in Fig.~\ref{fig:epipolar} (a). Given a point $\boldsymbol{p}$ on the image plane of the target view $i$ and the relative camera pose $\boldsymbol{\pi}^{i \to j}$, the goal is the compute the corresponding epipolar line $\boldsymbol{e}$ on the image plane of the source view $j$. We first compute the 3D ray of point $\boldsymbol{p}$, then we project any two 3D points on this ray onto the source view image plane. The line connected by these two projected points is the epipolar line $\boldsymbol{e}$. 

As mentioned in the main text, after getting the epipolr line, we compute pixel distances between patch centers from different views and invert it to weight map $K_{i, j}$. Finally, we directly add $K_{i, j}$ to the original attention weights $A_{i, j}$: $A^{\prime}_{i,j} = A_{i,j} + K_{i, j}$. In Fig.~\ref{fig:epipolar} (b), for the blue point in the left figure, we show its corresponding inverted epipolar distances on the right side.

\noindent
\textbf{UNet Block}. The UNet block consists of three modules: (i) a convolution layer; (ii) a self-attention layer among multiple output views; (iii) Cross-attention layer between input views and output views. The noisy latent images $\boldsymbol{y}$ (concatenated with $\tilde{\boldsymbol{x}}^{\mathrm{target}}$ predicted by SRT) are first passed to a convolution layer to obtain patch-level image features. Secondly, instead of performing self-attention within each individual view, we merge tokens from multiple output views and apply self-attention across all these tokens. Finally, we concatenate patch tokens with the corresponding ray embedding, and apply an epipolar attention between individual output views and input tokens (obtained by SRT).

\noindent
\textbf{Hyper-parameters}. Hyper-parameters used in our method are reported in Tab.~\ref{tab:hyper-params}.

\begin{table}[t]
\centering
\small
\setlength{\tabcolsep}{11pt}
\def\arraystretch{1}
\resizebox{\linewidth}{!}{
\scriptsize
\begin{tabular}{lc}
	\toprule
	Parameter & Value \\
        \cmidrule(lr){1-2} 
        Input view num & 1 - 4 \\
        Output view num & 1 - 4 \\
        Image size & 256 \\
        Latent image size (after vae) & 32 \\
        \cmidrule(lr){1-2} 
        SRT feature dim & 768 \\
        SRT head num & 12 \\
        SRT activation & gelu \\
        SRT-Enc layer num & 8 \\
        SRT-Dec layer num & 4 \\
        \cmidrule(lr){1-2} 
        UNet down/up block num & 4 \\
        UNet block out channels & (320, 640, 1280, 1280) \\
        UNet in channels & 8 \\
        UNet out channels & 4 \\
        UNet activation & silu \\
        \cmidrule(lr){1-2} 
        Diffusion steps (training) & 1000 \\
        Diffusion steps (inference) & 50 \\
        Sampling method & PNDM \\
        \cmidrule(lr){1-2} 
        Optimizier & AdamW \\
        Learning rate & $5e^{-5}$ \\
        AdamW Betas & (0.9, 0.999) \\
        Iterations & 100,000 \\
        Batch Size & 1024 \\
        Diffusion loss weight & 1.0 \\
        Reconstruction loss weight & 0.5 \\
	\bottomrule
\end{tabular}
}
\vspace{-5pt}
\caption{We report main hyper-parameters used in our method here.}
\label{tab:hyper-params}
\vspace{-15pt}
\end{table}

\section{Experiment Details}

\noindent
\textbf{Dataset}. We directly use the rendering data and the training/test split of the Objaverse dataset~\cite{deitke2023objaverse} provided by Zero123~\cite{DBLP:journals/corr/abs-2303-11328}. The whole dataset contains 800k objects, with 7k for testing. For each object, Zero123 provides 12 random views sampled from a uniform sphere. In addition, we employ the Google Scanned Object (GSO) dataset to evaluate the zero-shot ability of our method. For GSO, we directly use the rendering data provided by IBRNet~\cite{DBLP:conf/cvpr/WangWGSZBMSF21}. During evaluation, we take the first frame of each object as the source frame and employ the subsequent frame as the target frame.

\noindent
\textbf{Metrics}. We employ the VGG model, pre-trained on the ImageNet dataset, to calculate the LPIPS score. While computing the flow warping error $E_{\mathrm{warp}}$, the pre-trained RAFT~\cite{DBLP:conf/eccv/TeedD20} model is used to to compute the optical flow between consequent frames. For each object, a 32-frame video is generated where phi spans 360 degree, and the elevation is set to 10 degree with a radius of 1.5.

\noindent
\textbf{Reproducing Baselines}. 

\begin{itemize}
    \item Zero123. We use the official implementation on GitHub\footnote{\url{https://github.com/cvlab-columbia/zero123}} and the checkpoint pre-trained on Objaverse. We keep all the hyper-parameters and use their demo example to perform novel view synthesis. During inference, the guidance scale is set to 3.0 and the image is resized to $256 \times 256$.
    \item Point-E. We use the official implementation and pre-trained models on GitHub\footnote{\url{https://github.com/openai/point-e}}. We follow their example demo to perform 3D reconstruction.
    \item PoseDiff. While the code is publicly available at the moment, we faithfully re-implement their model based on the description in the paper. All hyper-parameters are consistent with our approach.
\end{itemize}

\end{document}